\documentclass[sn-basic]{sn-jnl}

\usepackage{algorithm}
\usepackage{tabularx}
\usepackage{multirow}
\usepackage{arydshln}

\jyear{2021}%

\theoremstyle{thmstyleone}%
\theoremstyle{thmstyletwo}%

\theoremstyle{thmstylethree}%

\DeclareMathAlphabet{\mathsb}{OT1}{cmr}{b}{n}
\newcommand{\vcsb}[1]{\ensuremath{\mathsb{#1}}} 
\DeclareMathOperator*{\argmax}{argmax}

\newcommand*{\boldtheta}{\boldsymbol{\theta}}

\newcommand*{\boldpsi}{\boldsymbol{\psi}}
\newcommand*{\boldphi}{\boldsymbol{\phi}}

\newcommand*{\boldtau}{\boldsymbol{\tau}}
\newcommand*{\boldgamma}{\boldsymbol{\gamma}}

\newcommand*{\support}{S_{i}}
\newcommand*{\query}{Q_{i}}
\newcommand*{\dataset}{\mathcal{D}_{i}}

\begin{document}

\title[Task-Adaptive Pseudo Labeling for Transductive Meta-Learning]{Task-Adaptive Pseudo Labeling for Transductive Meta-Learning}


\author{\fnm{Sanghyuk} \sur{Lee}}\email{xxxblanky@gmail.com}
\author{\fnm{Seunghyun} \sur{Lee}}\email{lsh910703@gmail.com}
\author*{\fnm{Byung~Cheol} \sur{Song}*}\email{bcsong@inha.ac.kr}

\affil{\orgdiv{Department of Electronic Engineering}, \orgname{Inha University}, \orgaddress{\city{Incheon}, \postcode{22212}, \country{South Korea}}}


\abstract{
    Meta-learning performs adaptation through a limited amount of support set, which may cause a sample bias problem. To solve this problem, transductive meta-learning is getting more and more attention, going beyond the conventional inductive learning perspective. This paper proposes so-called task-adaptive pseudo labeling for transductive meta-learning. Specifically, pseudo labels for unlabeled query sets are generated from labeled support sets through label propagation. Pseudo labels enable to adopt the supervised setting as it is and also use the unlabeled query set in the adaptation process. As a result, the proposed method is able to deal with more examples in the adaptation process than inductive ones, which can result in better classification performance of the model.
    Note that the proposed method is the first approach of applying task adaptation to pseudo labeling.
    Experiments show that the proposed method outperforms the state-of-the-art (SOTA) technique in 5-way 1-shot few-shot classification.
}




\maketitle

\section{Introduction}
    A major difference between machines and humans in the learning process is whether or not prior knowledge is used. Machines cannot apply the knowledge they learned in the past to the current task easily, but humans can do it quickly with little cost. So, many researchers have actively studied on few-shot learning~\cite{jiang2020multi,shao2021mdfm,zhou2022forward}, which tries to learn with little data. Especially, meta-learning, which learns meta-knowledge~\cite{survey} existing in task distribution and then transfers the result to a new target task, is receiving a lot of attention. Model-Agnostic Meta-Learning (MAML)~\cite{MAML} is the most famous one. MAML learns the initialization of model parameters with meta-knowledge to transfer. Note that this initialization contains task-generic knowledge, making it possible to adapt to a new task in fewer steps.

\begin{figure}[t]
    \begin{center}
        \includegraphics[width=.8\linewidth]{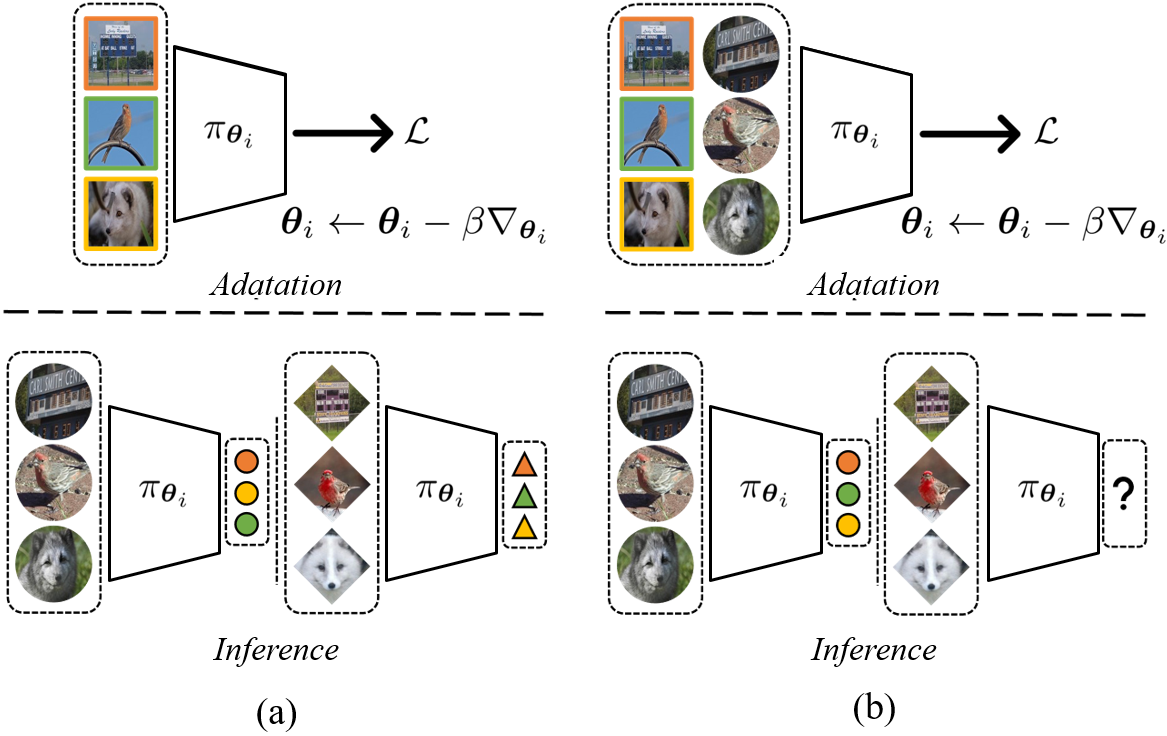}
    \end{center}
    \caption{Comparison of the (a) inductive and (b) transductive meta-learning. Transductive meta-learning utilizes an unlabeled query set (without edge color) in adaptation, while inductive meta-learning doesn't. Due to the limited number of samples in the support set, inductive meta-learning is vulnerable to suffering sample bias during adaptation.}
\label{fig1:concept}
\end{figure}
    
    A mini-batch of few-shot learning consists of tasks~\cite{MatchingNet,zhang2021learning}, rather than examples. A task in few-shot classification means classification of $N$ classes. The task, which is composed of a support set for adaptation and a query set for evaluating the adapted model, usually follows 1-shot or 5-shot settings. That is, a support set has 1 or 5 images per class, which means that adaptation to a target task can be optimized with only a very small number of data. On the contrary, the small data cannot sufficiently represent the corresponding class. This is called sample bias problem~\cite{sample_bias}.
    To solve the sample bias problem, \cite{TPN,SCA,MeTAL} applied transductive learning~\cite{transductive} to few-shot learning and its conceptual visualization is shown as Figure~\ref{fig1:concept}.
    Transductive learning has a premise that it can directly access test samples, while inductive learning trains a generalized model to work well on random test samples.
    That is, transductive learning in meta-learning has the effect of adaptation through a larger amount of data, by using the unlabeled query set together.
    Since the labels of the query set should not be used for training in general, a new methodology for calculating the loss from the unlabeled query set is required.
    Therefore, if unlabeled test samples are used for the learning process, transductive learning-based model will work better than inductive learning-based one at least for the target test samples, although the performance for the other samples may be slightly deteriorated.
    \begin{figure*}[t]
    \begin{center}
        \includegraphics[width=\linewidth]{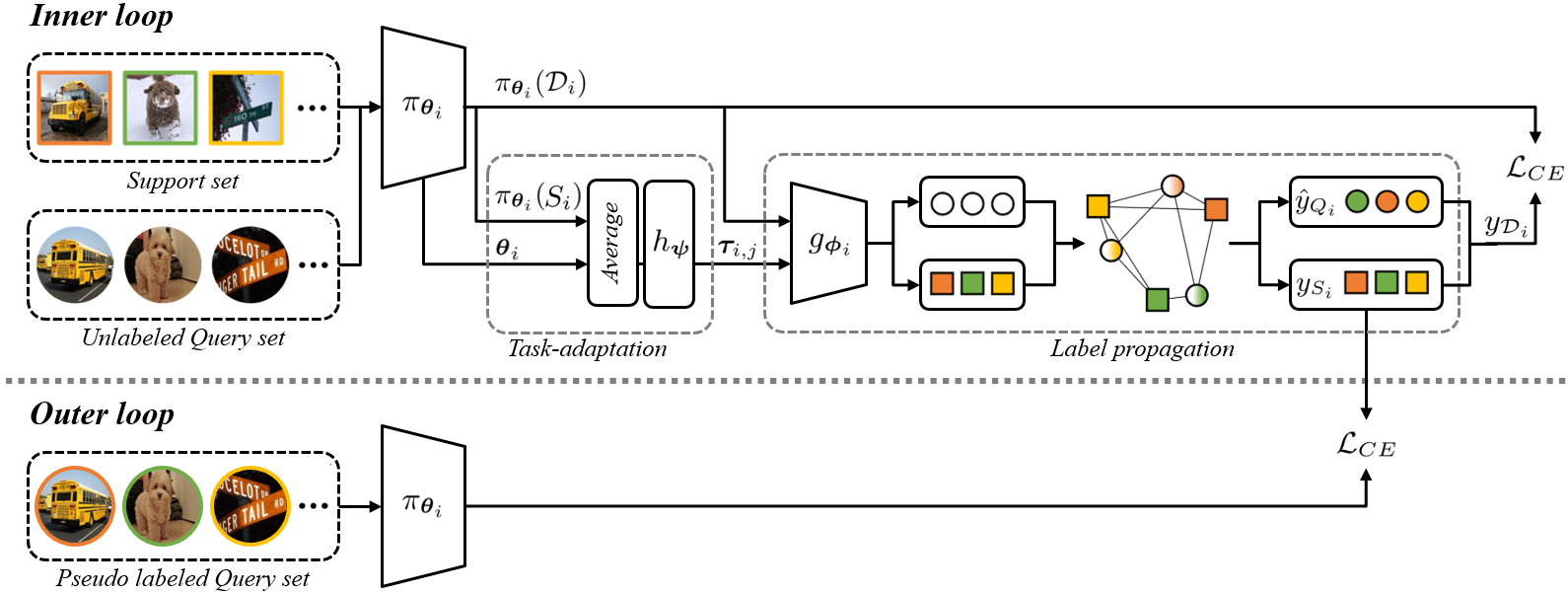}
    \end{center}
    \caption{
    Overview of the proposed method. We generate pseudo labels via label propagation for the unlabeled query set from the support set. 
    Here, a square and a circle denote the examples of the support set and the query set, respectively.
    Each color indicates a class of the current task.
}
\label{fig2:overall}
\end{figure*}
    
    This paper proposes Task-Adaptive Pseudo Labeling (TAPL) that is the first approach to apply task-adaptation to pseudo-labeling for transductive meta-learning. TAPL assigns pseudo labels on unlabeled query sets in an adaptive way to the target task.
    In detail, we adopt Transductive Propagation Network (TPN)~\cite{TPN}, which propagates the labels of the support sets to unlabeled query sets in a closed form using a graph neural network (GNN) during the inference process. TAPL delivers pseudo labels of unlabeled query sets from support sets via label propagation. Next, the pseudo-labeled query sets and support sets are adapted together through supervised settings as in MAML. Since the pseudo labels are hard labels, i.e., one-hot vector forms, the label propagation process cannot be fine-tuned for the target task. So, we present a small sub-network that generates layer-wise task-adaptive parameters for the graph construction network of TPN to propagate the task-adaptive labels. Since this sub-network receives a representation vector representing the current task, the weights of the graph construction network can be conditional to the task.
    
    The contribution points of the proposed method are summarized as follows:
    \begin{itemize}
        \item The proposed method proposes task-adaptive pseudo labeling for transductive meta-learning to mitigate the sample bias in adaptation, which didn’t draw lots of attention in meta-learning.
        \item The proposed method shows SOTA or comparable performance on the two famous datasets of few-shot classification purpose.
    \end{itemize}
    
    \subsection{Related work}\label{ssec:related}
        Meta-learning aims to transfer meta-knowledge, i.e., common information shared between tasks~\cite{learning_to_learn}. Here, well-known meta-learning strategies can be categorized into metric-based~\cite{siamese,ProtoNet}, optimization-based~\cite{L2L_by_GD,MAML,BGO}, model-based~\cite{EMM,MANN,MPM} algorithms. Specifically, model-based algorithms utilize a sub-network model to encode and parameterize the adaptation process, e.g., by encoding the training set~\cite{meta-networks}. Therefore, the proposed TAPL can be categorized as a model-based algorithm.
        
        In the view of the transductive meta-learning, TPN~\cite{TPN} proposed the label propagation from the support set for the inference of the query set. Even though TPN makes unlabeled query sets trainable via supervised learning, it still has a limitation in that the label propagation process is not task-adaptive. MeTAL~\cite{MeTAL} proposed task adaptive loss functions for neural networks. However, this approach has a disadvantage in that various factors (e.g., architecture, input type) must be carefully considered when designing its loss function.

\section{Proposed Method}
    \label{sec:method}
    In order to use the information of unlabeled query sets in the optimization process of adaptation, a kind of bypass method that does not utilize ground truth (GT) labels is required.
    So, conventional transductive meta-learning~\cite{SCA,MeTAL} defined label-free loss functions through neural networks.
    We were inspired from TPN~\cite{TPN}, which propagates labels from support set to query set.
    Specifically, TPN generates a graph using samples in both support set and unlabeled query set.
    Then, it applies a closed-form solution and gets prediction for the unlabeled query set.
    We came up with an idea of getting pseudo labels for the unlabeled query set by utilizing this label propagation method.
    Using this idea, we can utilize even unlabeled query sets in fully supervised setting during the adaptation process.
    
    On the other hand, we do not adopt TPN as it is for pseudo labeling because TPN cannot propagate labels adaptively to tasks changing in mini-batch units.
    In detail, since pseudo labels are trained in the form of hard labels and the argmax operation for extracting hard labels is non-differentiable, adaptation to the target task cannot be made in an end-to-end manner.
    
    To realize the effective pseudo-labeling for the target task, we propose to apply a task-adaptatation scheme to label propagation.
    One might question whether it would be better to utilize pseudo labels as soft labels to fully bring out the advantage of end-to-end learning.
    However, note that MAML requires a significant memory cost due to bi-level optimization.
    So, if label propagation using pseudo labels as soft labels is included in adaptation, more memory and computational cost may be required.
    On the other hand, using pseudo labels as hard labels (or one-hot vectors) is closely related to entropy minimization~\cite{entropy_min}.
    Therefore, we determine to use pseudo labels as hard labels in the adaptation process and present an additional module to make the weights of graph construction network conditional to the current task, which enables task adaptive pseudo labeling.

    \subsection{Task-Adaptive Pseudo Labeling}\label{ssec:TAPL}
        \begin{algorithm}[t]\small
\caption{Task-Adaptive Pseudo Labeling (TAPL)}

\makeatletter
\newcommand{\algmargin}{\the\ALG@thistlm}
\makeatother
\algnewcommand{\parState}[1]{\State%
  \parbox[t]{\dimexpr\linewidth-\algmargin}{\strut #1\strut}}

\textbf{Require: } $\beta, \eta$: Learning rates\\
\textbf{Require: } $J$: The number of inner-loop update steps

\begin{algorithmic}[1]
\State Randomly initialize $\boldtheta$, $\boldphi$, and $\boldpsi$

\While{not done}
    \State Sample $B$ numbers of mini-batch of the current task
    
    \For{$i \in [B]$}
        \State $\boldtheta_{i} \gets \boldtheta$
        \State $\boldphi_{i} \gets \boldphi$
        \State $\dataset = \{ \support, \query \}$
        
        \For {$j \in [J]$ }
            \State Compute $\pi_{\boldtheta_{i}}(\dataset)$
            \parState {%
                Get task representation $\boldtau_{i,j}$}
            \parState {%
                Generate task-adaptation parameters for $g_{\boldphi_{i}}$:\\
                $\{ \gamma^{\ell}_{i} \}_{\ell=1}^{L} = h_{\boldpsi}(\boldtau_{i,j})$}
            \parState {%
                Make the parameters $\boldphi_{i}$ adapted on the current task:\\
                $\{ \boldphi^{\ell}_{i} \}_{\ell=1}^{L} \gets \{ \gamma^{\ell}_{i} \boldphi^{\ell}_{i} \}_{\ell=1}^{L}$}
            \parState {%
                Generate pseudo labels for the unlabeled query sets:\\
                $\hat{y}_{\query}=\argmax{g_{\boldphi_{i}}(\pi_{\boldtheta_{i}}(\dataset))}$}
            \State Concatenate the labels: $y_{\dataset} = \{ y_{\support}, \hat{y}_{\query} \}$
            \parState {%
                Compute inner loop loss: $\mathcal{L}(\pi_{\boldtheta_{i}}(\dataset), y_{\dataset})$}
            \parState {%
                Perform gradient descent:\\
                $\boldtheta_{i} \gets \boldtheta_{i} - \beta \nabla_{\boldtheta_{i}} \mathcal{L}(\pi_{\boldtheta_{i}}(\dataset), y_{\dataset})$}
        \EndFor
    \parState{%
        Compute outer loop loss:\\
        $\mathcal{L}_{i} = \dfrac{1}{2}(\mathcal{L}(\pi_{\boldtheta_{i}}(\query), y_{\query}) + 
        \mathcal{L}(g_{\boldphi_{i}}(\pi_{\boldtheta_{i}}(\mathcal{D}_{i})), y_{\query}))$}
    \EndFor
    
    \parState {%
        Update the model parameters: \\
        $(\boldtheta, \boldphi, \boldpsi) \gets (\boldtheta, \boldphi, \boldpsi) - \eta \nabla_{(\boldtheta, \boldphi, \boldpsi)}\sum_{i=1}^{B} \mathcal{L}_{i}$
        }
    
\EndWhile

\end{algorithmic}
\label{alg:TAPL}
\end{algorithm}
        Figure~\ref{fig2:overall} shows the configuration and operation of the proposed method. First, we describe the proposed inner loop composed of task-adaptation and label propagation. Following~\cite{SCA, MeTAL}, each task can be represented by components of a neural network $\pi_{\theta}$, i.e., its parameters $\theta$ and logits $\pi_{\theta}(\cdot)$. The parameters are optimized by a given dataset, so they can represent a data distribution while extracting features. Meanwhile, logits intuitively represent a label space of the given task.
        We also construct a task represented vector $\boldtau$ in the same way. All the parameters and logits are averaged independently and concatenated, thereby composing a single feature vector. Assuming the few-shot classification is configured of $N$-way $K$-shot and the number of layers in the backbone network is $M$, the dimension of $\boldtau$ becomes $(NK + M)$.
        
        Next, we utilize $\boldtau$ to make the label propagation process task-adaptive. In order to train the unlabeled query set, labels of the support set are propagated to the query set through the graph construction network $g_{\boldphi_{i}}$, which will be explained later. Here, we modulate each parameter of $g_{\boldphi_{i}}$ using the task represent vector $\boldtau$. A small multi-layer perceptron $h_{\boldpsi}$, with $\boldtau$ as input, generates a task-adaptation parameter $\boldgamma_{i}$. The task-adatation process for the graph construction network is defined as follows:
        \begin{equation}
            \{ \phi^{\ell}_{i} \}^{L}_{\ell=1} \gets \{ \gamma^{\ell}_{i} \phi^{\ell}_{i} \}^{L}_{\ell=1}
        \end{equation}
        where $L$ is the number of layers of $g_{\boldphi_{i}}$. By doing so, $g_{\boldphi_{i}}$ becomes conditional to the current task, and we can expect the label propagation process to work more properly on the task. 
        If the complexity of the sub-network $h_{\boldpsi}$ is $C$, the complexity increase of the proposed method compared to TPN is $C(NK+M)$. Such a complexity increase is negligible because $h_{\boldpsi}$ uses a small MLP. The complexity issue is verified in the ablation study section.
        
        In the label propagation process, task-adapted $g_{\boldphi_{i}}$ takes logits both the support set and the query set and constructs a graph using them. The pseudo labels are first converted to hard labels $y_{\query}$ by applying $\argmax$ operation. With these pseudo labels, we are able to utilize the unlabeled query set for adapting to the target task in a fully supervised fashion. Next, by concatenating the hard labels and the label of support set $y_{\support}$, the label $y_{\dataset}$ for $\dataset$, which is a union of support set samples and query set samples, is generated. By using $y_{\dataset}$, adaptation for a given task is performed in the inner loop, and the parameter of the classifier $\theta$ is optimized.
        
        Finally, the outer loop utilizes the pseudo labeled query set and optimizes all the trainable parameters, i.e., $\theta$, $\boldpsi$, and $\boldphi$. However, as described above, if $y_{\query}$ is used as a hard label, $\boldphi_{i}$ is not learnable due to a non-differentiable operation, i.e., $\argmax$. So, we calculate the cross-entropy loss between the output of $g_{\boldphi_{i}}$ and the GT label of the query set prior to applying $\argmax$. Then, the outer loop can proceed to learn with respect to $\boldphi$, thereby can propagate more accurate pseudo labels.
        
        As a result, the model can avoid sample bias phenomenon in the inner loop because the model is optimized with more samples. Algorithm~\ref{alg:TAPL} summarizes the proposed method.
        
\section{Experiments}\label{sec:experiments}
    \begin{table*}[t]
\caption{Test accuracy of 5-way few-shot classification in miniImageNet and tieredImageNet. Here, `Trans.' indicates the tranductive learning}
\begin{center}
\fontsize{6}{6}\selectfont
\begin{tabularx}{\textwidth}{>{\arraybackslash}X
>{\centering\arraybackslash}>{\hsize=.3\hsize}X
>{\centering\arraybackslash}>{\hsize=.7\hsize}X
>{\centering\arraybackslash}>{\hsize=.7\hsize}X
>{\centering\arraybackslash}>{\hsize=.7\hsize}X
>{\centering\arraybackslash}>{\hsize=.7\hsize}X}
\toprule 
\multirow{2}{*}{\textbf{Method}} & 
\multirow{2}{*}{\textbf{Trans.}} & 
\multicolumn{2}{c}{\textbf{miniImageNet}} & 
\multicolumn{2}{c}{\textbf{tieredImageNet}} \\
& & \textbf{1-shot} & \textbf{5-shot} & \textbf{1-shot} & \textbf{5-shot} \\
\midrule
MAML~\cite{MAML}                &            &  $48.70 \pm 1.84\%$    & $63.11 \pm 0.92\%$     & $49.06 \pm 0.50\%$  & $67.48 \pm 0.47\%$\\
BOIL~\cite{BOIL}                &            &  $49.61 \pm 0.16\%$    & $66.45 \pm 0.37\%$     & $48.58 \pm 0.27\%$  & $69.37 \pm 0.12\%$\\
MAML++~\cite{MAML++}            &            &  $52.15 \pm 0.26\%$    & $68.32 \pm 0.44\%$     & --                  & --\\
ALFA~\cite{ALFA}                &            &  $50.58 \pm 0.51\%$    & $69.12 \pm 0.47\%$     & $53.16 \pm 0.49\%$  & $70.54 \pm 0.46\%$\\
L2F~\cite{L2F}                  &            &  $52.10 \pm 0.50\%$    & $69.38 \pm 0.46\%$     & $54.40 \pm 0.50\%$  & $73.34 \pm 0.44\%$\\
CxGrad~\cite{CxGrad}            &            & $51.80 \pm 0.46\%$    & $69.82 \pm 0.42\%$      & $55.55 \pm 0.46\%$  & $73.55 \pm 0.41\%$\\
\hdashline
TPN~\cite{TPN}                  & \checkmark & $53.75 \pm 0.86\%$    & $69.43 \pm 0.67\%$      & $57.53 \pm 0.96\%$  & $72.85 \pm 0.74\%$\\
TPN (Higher Shot)~\cite{TPN}    & \checkmark & $55.51 \pm 0.86\%$    & $69.86 \pm 0.65\%$      & $59.91 \pm 0.94\%$  & $73.30 \pm 0.75\%$\\
MAML++ + SCA~\cite{SCA}         & \checkmark & $54.24 \pm 0.99\%$    & $71.85 \pm 0.53\%$      & --    & --\\
MAML + MeTAL~\cite{MeTAL}       & \checkmark & $52.63 \pm 0.37\%$    & $70.52 \pm 0.29\%$      & $54.34 \pm 0.31\%$  & $70.40 \pm 0.21\%$\\
ALFA + MeTAL~\cite{MeTAL}       & \checkmark & $57.75 \pm 0.38\%$    & \vcsb{74.10 \pm 0.43\%} & $60.29 \pm 0.37\%$  & $75.88 \pm 0.29\%$\\
CxGrad + TAPL (Ours)            & \checkmark & \vcsb{58.55 \pm 0.46\%} & $72.28\pm0.41\%$    & \vcsb{60.58 \pm 0.45\%}    & $75.76 \pm 0.40\%$\\
\bottomrule
\end{tabularx}
\end{center}
\label{table1:benchmark}
\end{table*}
    In all experiments, we used four convolutional blocks from~\cite{MatchingNet}. Each convolutional block consists of one convolution layer with a kernel of size 3$\times$3, one batch normalization layer~\cite{batchnorm}, and one ReLU. A max-pooling layer with 2$\times$2 kernel and stride 2 is located between convolutional blocks. Adam~\cite{ADAM} was used in the outer loop for the optimization.
    The sub-network $h_{\boldpsi}$ is a simple 2-layer MLP.
    A ReLU activation function exists between the linear layers, and the sigmoid function is located at the last stage.
    
    We adopted two popular datasets of few-shot classification purpose. First, miniImageNet~\cite{optim_as_model} extracted from ImageNet~\cite{ImageNet} consists of 100 classes and 60,000 images. Each class contains 600 images of 84$\times$84. 100 classes are again divided into three folds; 64, 16, and 20 classes are used for meta-train, meta-validation, and meta-test, respectively, and the classes do not overlap each other. The second dataset, tieredImageNet~\cite{tieredImageNet}, is a larger one than miniImageNet. In a similar way to miniImageNet, 608 classes were extracted from ImageNet and composed of 779,165 images. The image resolution is the same as that of miniImageNet, but the classes are grouped into 34 upper categories unlike miniImageNet. That is, 20, 6, and 8 categories are used for meta-train, meta-validation, and meta-test, respectively, and these categories also do not overlap each other. Since classes are phase-wise mutually exclusive, both datasets are suitable for evaluating generalization ability.
    
    Note that since TAPL is orthogonal to conventional gradient-based meta-learning techniques, it can be used in a plug-in-play way. Therefore, we adopted a model combining TAPL and CxGrad~\cite{CxGrad}, which induced representation change by promoting backbone learning and ultimately improved the performance of MAML, for all the experiments in this paper.
    
    \subsection{Few-Shot Classification}
        \label{ssec:FSC}
        \begin{figure}[t]
\begin{center}
    \includegraphics[width=.8\linewidth]{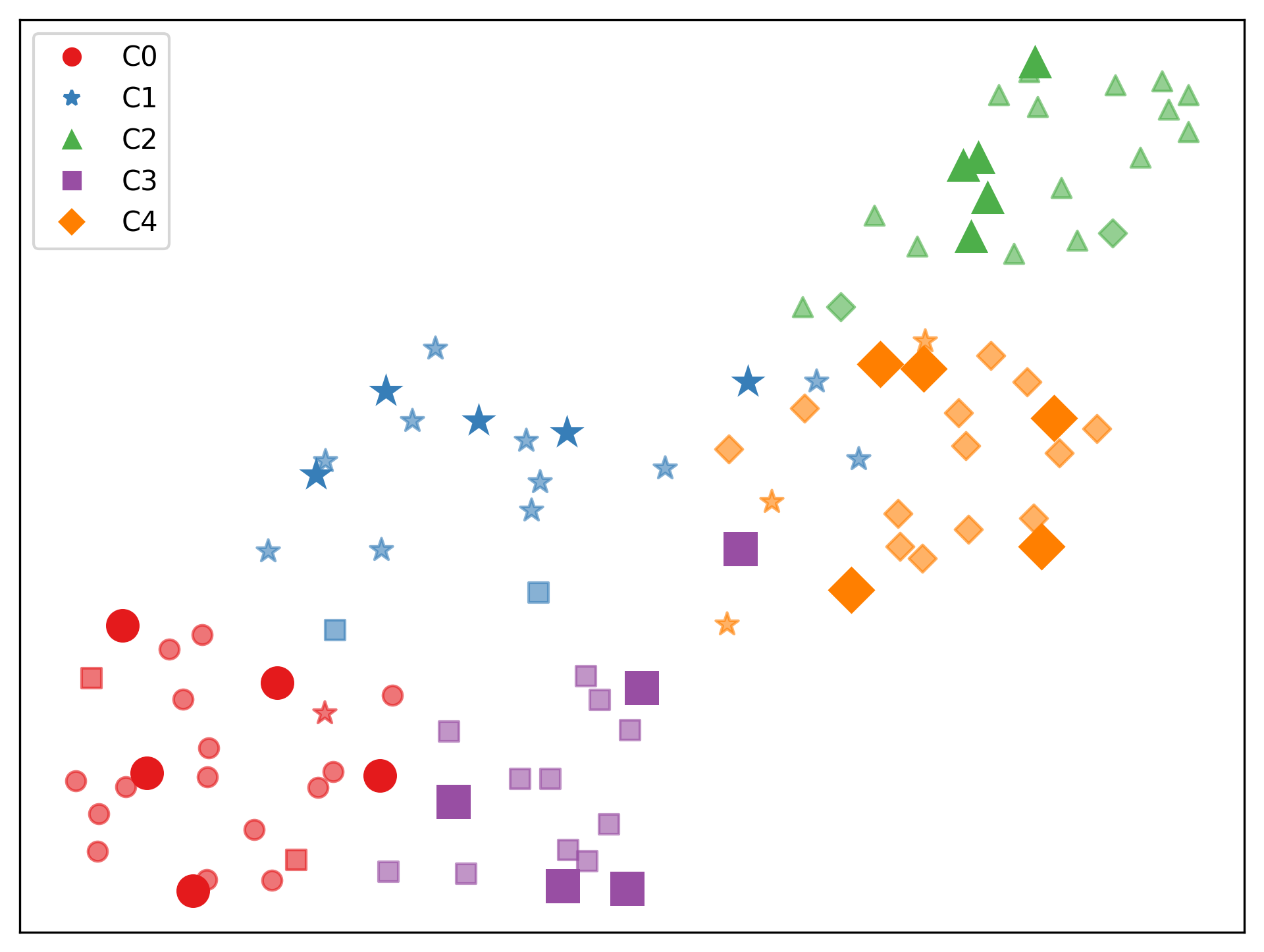}
\end{center}
\caption{Visualization of how pseudo labels are given to the unlabeled query sets in embedding space. Many unlabeled query set samples are given the same label as their ground truth one. Furthermore, the samples with the same pseudo label are embedded closely to each other.}
\label{fig3:visualization}
\end{figure}
        Table~\ref{table1:benchmark} shows the experimental result for 5-way 1-shot and 5-shot few-shot classification. Each result is from the ensemble of top 5 high accuracy cases in validation. In the case of 5-way 1-shot, the proposed method shows the SOTA performance for both datasets. For 5-shot, the proposed method provides a noticeable performance improvement of about 1.8\% over the third place in miniImageNet, even though its performance improvement is marginal in tieredImageNet.
        The reason the performance increase of 1-shot and 5-shot is noticeably different is due to the characteristic of the backbone framework, i.e., TPN. To propagate labels, TPN utilizes a simple GNN with small receptive fields, so it is hard to capture the information for 5-shot data. Therefore, a better GNN is required for many-shot data.
    
    \subsection{Visualization}
        \label{ssec:visualization}
        This section examines the relationship between the unlabeled query set and the support set in the embedding space through visualization. UMAP~\cite{UMAP} is a well-known non-linear dimension reduction method, and is widely used to visualize how examples are embedded. Using UAMP, we can find out what kind of relationship the unlabeled query set given with pseudo labels has with the support set of the same class.
        
        Figure~\ref{fig3:visualization} shows the UMAP for 5-way 5-shot in miniImageNet. The plane figures distinguish different GT classes, and each color represents the class pointed by the pseudo label. Figures with larger size and darker color correspond to support set, and the others with smaller size and lighter color corresponds to query set. For example, A sample in a query set with GT label C0 and pseudo label C1 is expressed by the blue color of C1 and the circle shape of C0. Figure~\ref{fig3:visualization} shows that most samples of the unlabeled query set are embedded close to the support set corresponding to each GT class. Also, the pseudo label is the same as the GT one. Therefore, the proposed method accomplishes effective pseudo labeling by constructing graphs adaptive to the target task, and consequently gives a positive effect on adaptation.
    
    \subsection{Ablation Study}
        \label{ssec:Ablation}
        \begin{table}[t]
\caption{Ablation study in task-adaptiveness and learnability of $\alpha$ in label propagation.}
\begin{center}
\begin{tabular}{cccc}
\toprule
\textbf{Task-Adaptive} & \textbf{Learnable $\alpha$} & \textbf{iteration time} & \textbf{Accuracy}\\
\midrule
           & \checkmark & 0.709s & $71.22 \pm 0.42\%$ \\
\checkmark & \checkmark & 0.719s & \vcsb{72.28 \pm 0.41\%} \\
\bottomrule
\end{tabular}
\end{center}
\label{table:ablation}
\end{table}
        We examined how to construct graphs adaptive to the target task, and then analyzed the parameter $\alpha \in (0,1)$ that controls the amount of propagated information.
        Also, we analyzed the complexity increase due to the sub-network for task adaptation. Training iteration time was measured on Quadro RTX 8000, The experimental results are shown in Table~\ref{table:ablation}. The larger $\alpha$, the greater the influence of the predicted label obtained from the graph structure. When task-adaptive graph construction and learnable $\alpha$ were used together, about 1.1\% improved over task-generic pseudo labeling. Therefore, TAPL plays a crucial role in improving the overall performance of the proposed method.
        Also, the training iteration time of the proposed method is 0.719s which increases by only 1.4\% from TPN. Therefore, we claim that TAPL has a very reasonable cost.

\section{Conclusion}
    \label{sec:conclusion}
    Transductive meta-learning uses an unlabeled query set in the adaptation process to solve the sample bias problem occurring in conventional meta-learning due to a limited amount of support set. This paper proposes a novel transductive meta-learning utilizing task-adaptive pseudo labeling. Specifically, pseudo labeling is performed on unlabeled query sets through label propagation. This allows the existing cross-entropy loss function to be used as it is, without a well-designed label-free loss function in the adaptation process. That is, it became possible to learn while maintaining the existing supervised setting. Extensive experiments showed that the proposed method accomplished effective adaptation, especially, achieved SOTA performance for 1-shot classification. However, due to a structural limitation, the 5-shot accuracy is not acceptable yet, so we remain this problem as a future work.

\bibliography{sn-bibliography}

\begin{thebibliography}{33}
\providecommand{\natexlab}[1]{#1}
\providecommand{\url}[1]{{#1}}
\providecommand{\urlprefix}{URL }
\providecommand{\doi}[1]{\url{https://doi.org/#1}}
\providecommand{\eprint}[2][]{\url{#2}}
 \bibcommenthead

\bibitem[{Andrychowicz et~al(2016)Andrychowicz, Denil, Gomez, Hoffman, Pfau,
  Schaul, Shillingford, and De~Freitas}]{L2L_by_GD}
Andrychowicz M, Denil M, Gomez S, et~al (2016) Learning to learn by gradient
  descent by gradient descent. Advances in neural information processing
  systems 29

\bibitem[{Antoniou and Storkey(2019)}]{SCA}
Antoniou A, Storkey AJ (2019) Learning to learn by self-critique. Advances in
  Neural Information Processing Systems 32

\bibitem[{Antoniou et~al(2019)Antoniou, Edwards, and Storkey}]{MAML++}
Antoniou A, Edwards H, Storkey A (2019) How to train your maml. In: Seventh
  International Conference on Learning Representations

\bibitem[{Baik et~al(2020{\natexlab{a}})Baik, Choi, Choi, Kim, and Lee}]{ALFA}
Baik S, Choi M, Choi J, et~al (2020{\natexlab{a}}) Meta-learning with adaptive
  hyperparameters. Advances in Neural Information Processing Systems
  33:20,755--20,765

\bibitem[{Baik et~al(2020{\natexlab{b}})Baik, Hong, and Lee}]{L2F}
Baik S, Hong S, Lee KM (2020{\natexlab{b}}) Learning to forget for
  meta-learning. In: Proceedings of the IEEE/CVF Conference on Computer Vision
  and Pattern Recognition, pp 2379--2387

\bibitem[{Baik et~al(2021)Baik, Choi, Kim, Cho, Min, and Lee}]{MeTAL}
Baik S, Choi J, Kim H, et~al (2021) Meta-learning with task-adaptive loss
  function for few-shot learning. In: Proceedings of the IEEE/CVF International
  Conference on Computer Vision, pp 9465--9474

\bibitem[{Cui and Guo(2021)}]{sample_bias}
Cui W, Guo Y (2021) Parameterless transductive feature re-representation for
  few-shot learning. In: International Conference on Machine Learning, PMLR, pp
  2212--2221

\bibitem[{Finn et~al(2017)Finn, Abbeel, and Levine}]{MAML}
Finn C, Abbeel P, Levine S (2017) Model-agnostic meta-learning for fast
  adaptation of deep networks. In: International conference on machine
  learning, PMLR, pp 1126--1135

\bibitem[{Grandvalet and Bengio(2004)}]{entropy_min}
Grandvalet Y, Bengio Y (2004) Semi-supervised learning by entropy minimization.
  Advances in neural information processing systems 17

\bibitem[{Hospedales et~al(2021)Hospedales, Antoniou, Micaelli, and
  Storkey}]{survey}
Hospedales T, Antoniou A, Micaelli P, et~al (2021) Meta-learning in neural
  networks: A survey. IEEE transactions on pattern analysis and machine
  intelligence 44(9):5149--5169

\bibitem[{Ioffe and Szegedy(2015)}]{batchnorm}
Ioffe S, Szegedy C (2015) Batch normalization: Accelerating deep network
  training by reducing internal covariate shift. In: International conference
  on machine learning, PMLR, pp 448--456

\bibitem[{Jiang et~al(2020)Jiang, Huang, Geng, and Deng}]{jiang2020multi}
Jiang W, Huang K, Geng J, et~al (2020) Multi-scale metric learning for few-shot
  learning. IEEE Transactions on Circuits and Systems for Video Technology
  31(3):1091--1102

\bibitem[{Kingma and Ba(2015)}]{ADAM}
Kingma DP, Ba J (2015) Adam: A method for stochastic optimization. In: ICLR
  (Poster)

\bibitem[{Koch et~al(2015)Koch, Zemel, Salakhutdinov et~al}]{siamese}
Koch G, Zemel R, Salakhutdinov R, et~al (2015) Siamese neural networks for
  one-shot image recognition. In: ICML deep learning workshop, Lille, p~0

\bibitem[{Lai et~al(2020)Lai, Kan, Han, Song, and Shan}]{MPM}
Lai N, Kan M, Han C, et~al (2020) Learning to learn adaptive
  classifier-predictor for few-shot learning. IEEE transactions on neural
  networks and learning systems 32(8):3458--3470

\bibitem[{Lee et~al(2022)Lee, Lee, and Song}]{CxGrad}
Lee S, Lee S, Song BC (2022) Contextual gradient scaling for few-shot learning.
  In: Proceedings of the IEEE/CVF Winter Conference on Applications of Computer
  Vision, pp 834--843

\bibitem[{Liu et~al(2018)Liu, Lee, Park, Kim, Yang, Hwang, and Yang}]{TPN}
Liu Y, Lee J, Park M, et~al (2018) Learning to propagate labels: Transductive
  propagation network for few-shot learning. arXiv preprint arXiv:180510002

\bibitem[{McInnes et~al(2018)McInnes, Healy, and Melville}]{UMAP}
McInnes L, Healy J, Melville J (2018) Umap: Uniform manifold approximation and
  projection for dimension reduction. arXiv preprint arXiv:180203426

\bibitem[{Munkhdalai and Yu(2017)}]{meta-networks}
Munkhdalai T, Yu H (2017) Meta networks. In: International Conference on
  Machine Learning, PMLR, pp 2554--2563

\bibitem[{Oh et~al(2020)Oh, Yoo, Kim, and Yun}]{BOIL}
Oh J, Yoo H, Kim C, et~al (2020) Boil: Towards representation change for
  few-shot learning. arXiv preprint arXiv:200808882

\bibitem[{Ravi and Larochelle(2017)}]{optim_as_model}
Ravi S, Larochelle H (2017) Optimization as a model for few-shot learning. In:
  International Conference on Learning Representations,
  \urlprefix\url{https://openreview.net/forum?id=rJY0-Kcll}

\bibitem[{Ren et~al(2018)Ren, Ravi, Triantafillou, Snell, Swersky, Tenenbaum,
  Larochelle, and Zemel}]{tieredImageNet}
Ren M, Ravi S, Triantafillou E, et~al (2018) Meta-learning for semi-supervised
  few-shot classification. In: International Conference on Learning
  Representations, \urlprefix\url{https://openreview.net/forum?id=HJcSzz-CZ}

\bibitem[{Russakovsky et~al(2015)Russakovsky, Deng, Su, Krause, Satheesh, Ma,
  Huang, Karpathy, Khosla, Bernstein et~al}]{ImageNet}
Russakovsky O, Deng J, Su H, et~al (2015) Imagenet large scale visual
  recognition challenge. International journal of computer vision
  115(3):211--252

\bibitem[{Santoro et~al(2016)Santoro, Bartunov, Botvinick, Wierstra, and
  Lillicrap}]{MANN}
Santoro A, Bartunov S, Botvinick M, et~al (2016) Meta-learning with
  memory-augmented neural networks. In: International conference on machine
  learning, PMLR, pp 1842--1850

\bibitem[{Shao et~al(2021)Shao, Xing, Xu, Liu, Wang, and Liu}]{shao2021mdfm}
Shao S, Xing L, Xu R, et~al (2021) Mdfm: Multi-decision fusing model for
  few-shot learning. IEEE Transactions on Circuits and Systems for Video
  Technology

\bibitem[{Snell et~al(2017)Snell, Swersky, and Zemel}]{ProtoNet}
Snell J, Swersky K, Zemel R (2017) Prototypical networks for few-shot learning

\bibitem[{Thrun and Pratt(1998)}]{learning_to_learn}
Thrun S, Pratt L (1998) Learning to learn: Introduction and overview. In:
  Learning to learn. Springer, p 3--17

\bibitem[{Vapnik(2006)}]{transductive}
Vapnik V (2006) 24 transductive inference and semi-supervised learning.
  Semi-Supervised Learning pp 453--472

\bibitem[{Vinyals et~al(2016)Vinyals, Blundell, Lillicrap, Wierstra
  et~al}]{MatchingNet}
Vinyals O, Blundell C, Lillicrap T, et~al (2016) Matching networks for one shot
  learning. Advances in neural information processing systems 29

\bibitem[{Zhang et~al(2021{\natexlab{a}})Zhang, Yuan, Zheng, Krikidis, and
  Wong}]{EMM}
Zhang J, Yuan Y, Zheng G, et~al (2021{\natexlab{a}}) Embedding model-based fast
  meta learning for downlink beamforming adaptation. IEEE Transactions on
  Wireless Communications 21(1):149--162

\bibitem[{Zhang et~al(2021{\natexlab{b}})Zhang, Zuo, Du, and
  Zhen}]{zhang2021learning}
Zhang L, Zuo L, Du Y, et~al (2021{\natexlab{b}}) Learning to adapt with memory
  for probabilistic few-shot learning. IEEE Transactions on Circuits and
  Systems for Video Technology 31(11):4283--4292

\bibitem[{Zhou et~al(2022)Zhou, Wang, Ye, Ma, Pu, and Zhan}]{zhou2022forward}
Zhou DW, Wang FY, Ye HJ, et~al (2022) Forward compatible few-shot
  class-incremental learning. In: Proceedings of the IEEE/CVF Conference on
  Computer Vision and Pattern Recognition, pp 9046--9056

\bibitem[{Zhu et~al(2020)Zhu, Li, Wu, Zhao, Ding, and Shi}]{BGO}
Zhu H, Li L, Wu J, et~al (2020) Personalized image aesthetics assessment via
  meta-learning with bilevel gradient optimization. IEEE Transactions on
  Cybernetics

\end{thebibliography}


\end{document}